\documentclass[11pt,a4paper]{article}
\usepackage[utf8]{inputenc}
\PassOptionsToPackage{hyphens}{url}
\usepackage[hyperref]{acl2017}
\usepackage{newtxtext}
\usepackage{newtxmath}
\usepackage{microtype}
\usepackage{latexsym}
\usepackage{booktabs}
\usepackage{tabularx}
\usepackage{graphicx}
\usepackage{multirow}
\usepackage{soul}
\usepackage{subfigure}
\usepackage{colortbl}
\usepackage{adjustbox}
\newcommand\model[1]{\textsf{#1}}

\aclfinalcopy 

\hypersetup{
pdftitle={Cross-topic Argument Mining from Heterogeneous Sources Using Attention-based Neural Networks},
pdfauthor={Christian Stab and Tristan Miller},
}

\title{Cross-topic Argument Mining from Heterogeneous Sources Using Attention-based Neural Networks}

\author{Christian Stab, Tristan Miller, Iryna Gurevych\\
Ubiquitous Knowledge Processing Lab (UKP-TUDA), \\
Department of Computer Science, Technische Universität Darmstadt\\
{\tt www.ukp.tu-darmstadt.de}}

\begin{document}
\maketitle

\begin{abstract}
Argument mining is a core technology for automating argument search in large document collections. Despite its usefulness for this task, most current approaches to argument mining are designed for use only with specific text types and fall short when applied to heterogeneous texts. In this paper, we propose a new sentential annotation scheme that is reliably applicable by crowd workers to arbitrary Web texts. We source annotations for over 25,000 instances covering eight controversial topics. The results of cross-topic experiments show that our attention-based neural network generalizes best to unseen topics and outperforms vanilla BiLSTM models by 6\% in accuracy and 11\% in F-score.
\end{abstract}

%------------------------------------------------
% INTRODUCTION
%------------------------------------------------
\section{Introduction}

Information retrieval and question answering are by now mature technologies that excel at answering factual queries on uncontroversial topics.  However, they provide no specialized support for queries where there is no single canonical answer, as with topics that are controversial or opinion-based.  For such queries, the user may need to carefully assess the stance, source, and supportability for each of the answers.  These processes can be supported by argument mining (AM), a nascent area of natural language processing concerned with the automatic recognition and interpretation of arguments.

In this paper, we apply AM to the task of \emph{argument search}---that is, searching a large document collection for arguments relevant to a given topic.  Searching for and classifying relevant arguments plays an important role in decision making~\cite{svenson1979process}, legal reasoning~\cite{wyner2010approaches}, and the critical reading, writing, and summarization of persuasive texts~\cite{kobayashi2009comprehension,wingate2012argument}.  Automating the argument search process could ease much of the manual effort involved in these tasks, particularly if it can be made to robustly handle arguments from different text types and topics.

But despite its obvious usefulness, this sort of argument search has attracted relatively little attention in the research community.  This may be due in part to the limitations of the underlying models and training resources, particularly as they relate to heterogeneous sources.  That is, most current approaches to AM are designed for use with particular text types and do not work well when applied to different data sets~\cite{daxenberger-EtAl:2017:EMNLP2017}.  Indeed, as \newcite{Habernal2014} observe, there is a great diversity of perspectives on how arguments can be best characterized and modelled, and no ``one-size-fits-all'' argumentation theory that applies to the variety of text sources found on the Web.

In this paper, we propose an argument annotation scheme that is (1)~applicable to the information-seeking perspective of argument search, (2)~general enough for use on heterogeneous data sources, and (3)~simple enough to be applied manually by untrained annotators.  We investigate whether it is possible to achieve reasonable data quality using crowdsourced annotations, and how well computational models trained on this data perform on the argument search task within and across different topics. Finally, we measure the amount of topic-specific data that must be added to a topic-general model in order for it to achieve in-topic performance comparable to that of a topic-specific model.

\begin{table*}
\centering
\footnotesize
  \begin{tabular}{l | p{100mm} | l}
  \toprule
    topic & sentence & label \\
  \midrule
	nuclear energy 	& Nuclear fission is the process that is used in nuclear reactors to produce energy using element called uranium.
 & no argument\\
   \midrule
 	nuclear energy	& It has been determined that the amount of greenhouse gases have decreased by almost half because of the prevalence in the utilization of nuclear power. 	& supporting argument \\
  	\midrule
  	minimum wage		& A 2014 study [\dots] found that minimum wage workers are more likely to report poor health, suffer from chronic diseases, and be unable to afford balanced meals. & opposing argument \\
 	\midrule
 	minimum wage		& We should abolish all Federal wage standards and allow states and localities to set their own minimums. & no argument \\
  \bottomrule
  \end{tabular}
  \caption{Example annotations illustrating our annotation scheme.} 
  \label{tab:examples}
\end{table*}

Our results show that crowd workers can indeed apply our annotation scheme to arbitrary Web texts quickly and reliably, allowing us to obtain huge amounts of data at a reasonable cost.  The corpus we produce includes over 25,000 instances over eight controversial topics, allowing for cross-topic experiments using heterogeneous text types.  The results of these experiments show that our attention-based neural network outperforms vanilla BiLSTM models in cross-topic experiments, with a relative improvement of 6\% in accuracy and 11\% in F-score. 

%------------------------------------------------
% RELATED WORK
%------------------------------------------------

\section{Related Work}

Most existing approaches treat argument mining at the discourse level, focusing on tasks such as segmenting argumentative discourse units~\cite{ajjour-EtAl:2017:ArgumentMining,Goudas2014}, classifying the function of argumentative discourse units (for example, as \emph{claims} or \emph{premises})~\cite{MochalesPalau2009,stab-gurevych:2014:EMNLP2014}, and recognizing argumentative discourse relations~\cite{eger-daxenberger-gurevych:2017:Long,Stab2017,nguyen-litman:2016:P16-1}.  These discourse-level approaches address the identification of argumentative structures within a single document but do not consider relevance to externally defined topics.

To date, there has been little research on the identification of topic-relevant arguments for argument search.  \newcite{wachsmuth-EtAl:2017:ArgumentMining} present a generic argument search framework. The system, however, relies on already structured arguments from debate portals and is not yet able to retrieve arguments from arbitrary texts.  \newcite{levy-EtAl:2014:Coling} investigate the identification of topic-relevant claims, an approach that was later extended with evidence extraction to mine supporting statements for claims~\cite{rinott-EtAl:2015:EMNLP}.  However, both approaches are designed to mine arguments from Wikipedia articles; it is unclear whether their annotation scheme is applicable to other text types or whether it can be easily and accurately applied by untrained annotators.  \newcite{hua-wang:2017:Short} identify sentences in cited documents that have been used by an editor to formulate an argument.  In contrast to this work, we do not limit our approach to the identification of sentences related to a given \emph{argument}, but rather focus on the retrieval of any argument relevant to a given \emph{topic}.  The fact that we are concerned with retrieval of arguments also sets our work apart from the discourse-agnostic stance detection task of \newcite{mohammad2016semeval}, which is concerned with the identification of sentences expressing support or opposition to a given topic, irrespective of whether those sentences contain supporting evidence (as opposed to mere statements of opinion).

Cross-domain AM experiments have so far been conducted only for discourse-level tasks such as claim identification~\cite{daxenberger-EtAl:2017:EMNLP2017}, argumentative segment identification~\cite{alkhatib-EtAl:2016:N16-1}, and argumentative unit segmentation~\cite{ajjour-EtAl:2017:ArgumentMining}.  However, the discourse-level argumentation models employed for these studies seem to be highly dependent on the text types for which they were designed; they do not work well when applied to other text types~\cite{daxenberger-EtAl:2017:EMNLP2017}.  The crucial difference between our own work and prior cross-domain experiments is that we investigate AM from heterogeneous texts across different \emph{topics} instead of studying specific discourse-level AM tasks across restricted text types of existing corpora.

%------------------------------------------------
% CORPUS CREATION
%------------------------------------------------

\section{Annotation Scheme and Corpus Creation}

There exists a great diversity in models of argumentation, which differ in their perspective, complexity, terminology, and intended applications~\cite{bentahar2010taxonomy}.  For the present study, we propose a model which, though simplistic, is nonetheless well-suited to the argument search scenario outlined in our introduction.  We define an \emph{argument} as a span of text expressing evidence or reasoning that can be used to either support or oppose a given topic.   An argument need not be ``direct'' or self-contained---it may presuppose some common or domain knowledge, or the application of commonsense reasoning---but it must be unambiguous in its orientation to the topic.  A \emph{topic}, in turn, is some matter of controversy for which there is an obvious polarity to the possible outcomes---that is, a question of being either \emph{for} or \emph{against} the use or adoption of something, the commitment to some course of action, etc. In some graph-based models of  argumentation~\cite[Ch.\,2]{stab2017argumentative}, what we refer to as a \emph{topic} would be part of a \emph{(major) claim} expressing a positive or negative stance, and our \emph{arguments} would be \emph{premises} with supporting\slash attacking \emph{consequence relations} to the claim.  However, unlike these models, which are typically used to represent (potentially deep or complex) argument structures at the discourse level, ours is a flat model that considers arguments in isolation from their surrounding context.  A great advantage of this approach is that it allows annotators to classify text spans without having to read large amounts of text and without having to consider relations to other topics or arguments.

In this work, we restrict ourselves to topics that can be concisely and implicitly expressed through keywords, and arguments that consist of individual sentences.  Some examples, drawn from our data set, are shown in Table~\ref{tab:examples}.  
The first three examples should be self-explanatory. The fourth example expresses opposition to the topic, but under our definition it is properly classified as a non-argument because it is a mere statement of stance that provides no evidence or reasoning.

\subsection{Data}\label{sec:expert_annotations}

For our experiments it was necessary to gather a large collection of manually annotated arguments that cover a variety of topics and that come from a variety of text types.  We started by randomly selecting eight topics (see Table~\ref{tab:corpus_stats}) from online lists of controversial topics.\footnote{\url{https://www.questia.com/library/controversial-topics}, \url{https://www.procon.org/}}  
For each topic, we made a Google query for the topic name, removed results not cached by the Wayback Machine,\footnote{\url{https://web.archive.org/}} and truncated the list to the top 50 results.  
This resulted in a set of persistent, topic-relevant, largely (but not exclusively) polemical Web documents representing a range of genres and text types, including news reports, editorials, blogs, debate forums, and encyclopedia articles.  We preprocessed each document with Apache Tika\footnote{\url{https://tika.apache.org/}} to remove boilerplate text. We then used the Stanford CoreNLP tools~\cite{manning2014stanford} to perform tokenization, sentence segmentation, and part-of-speech tagging on the remaining text, and removed all sentences without verbs or with less than three tokens.  This left us with a raw data set of 27,520 sentences (about 2,700 to 4,400 sentences per topic).

\begin{figure}
  \centering{
    \includegraphics[width=1.0\linewidth]{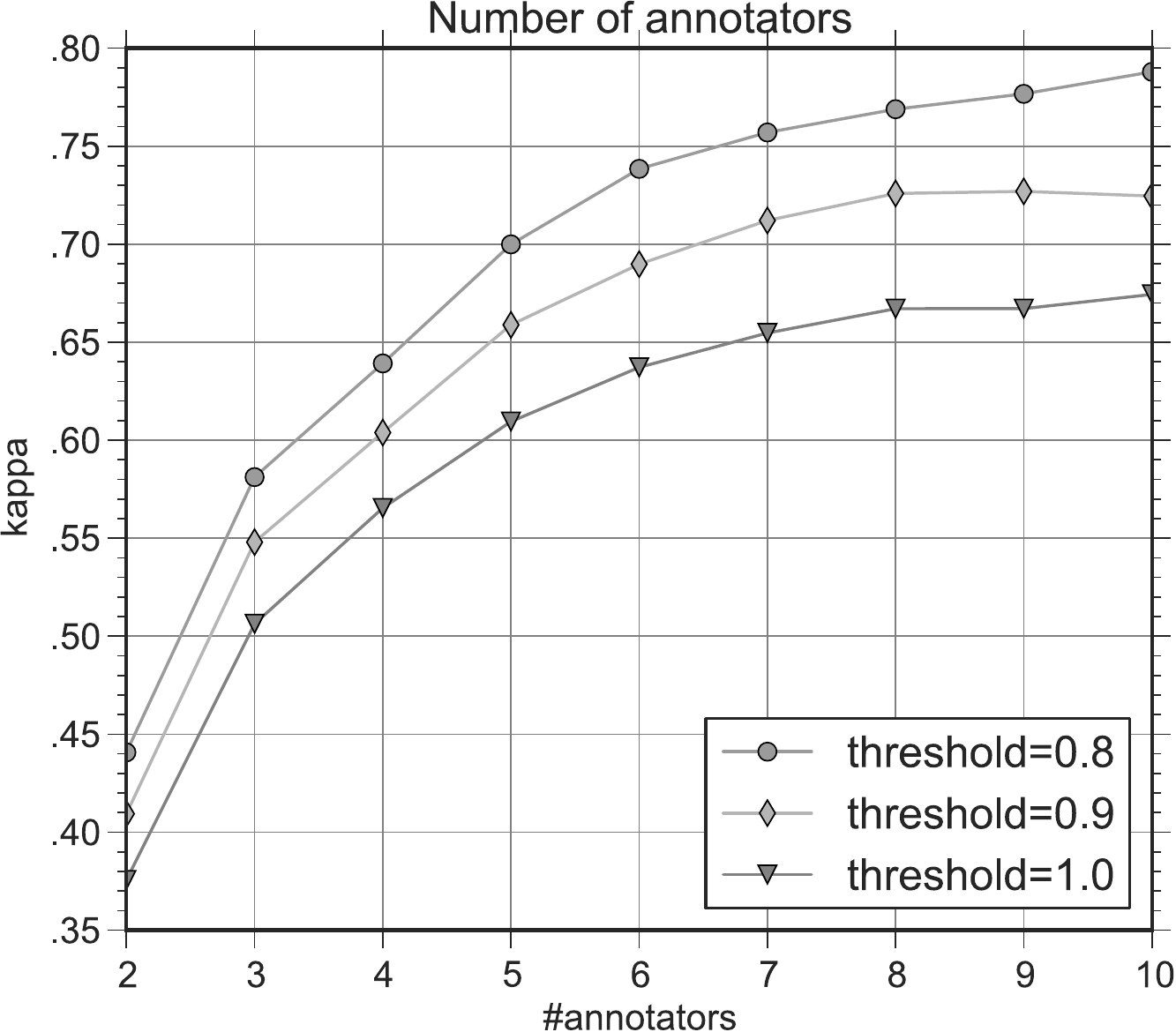}
    \caption{Influence of the number of crowd annotators and different MACE thresholds on $\kappa$.}
    \label{fig:num_annotators}
  }
\end{figure}

To assist annotators in classifying these sentences according to our argumentation model, we created a browser-based annotation interface that presents a brief set of instructions, a topic, a list of sentences, and a multiple-choice form for specifying whether each sentence is a supporting argument, an opposing argument, or not an argument with respect to the topic.  

\subsection{Analysis}
\label{sec:analysis}

To test the applicability of our annotation scheme by untrained annotators, we performed an experiment where we had a group of expert annotators and a group of untrained annotators classify the same set of sentences, and then compared the two groups' classifications.  The data for this experiment consisted of 200 sentences randomly selected from each of our eight topics.  Our ``expert'' annotators were two graduate-level language technology researchers who were fully briefed on the nature and purpose of the argument model.  Our untrained annotators were anonymous American workers from the Amazon Mechanical Turk (AMT) crowdsourcing platform.  Each sentence was independently annotated by the two expert annotators and ten crowd workers.

Inter-annotator agreement for our two experts, as measured by Cohen's $\kappa$, was 0.721; this exceeds the commonly used threshold of 0.7 for assuming the results are reliable~\cite{carletta1996assessing}.  We proceeded by having the two experts resolve their disagreements, resulting in a set of ``expert'' gold-standard annotations.  Similar gold standards were produced for the crowd annotations by applying the MACE de-noising tool~\cite{hovy2013learning}; we tested various threshold values (1.0, 0.9, and 0.8) to discard instances that cannot be confidently assigned a canonical label.  We then calculated Cohen's $\kappa$ between the remaining instances in the expert and crowd gold standards.  In order to determine the relationship between inter-annotator agreement and the number of crowd workers, we performed this procedure with successively lower numbers of crowd workers, going from the original ten annotators per instance down to two.  The results are visualized in Figure~\ref{fig:num_annotators}.  We observe that using seven annotators and a MACE threshold of 0.9 results in $\kappa = 0.723$; this gives us similar reliability as with the expert annotators without sacrificing too much coverage.  Table~\ref{tab:agreement} shows the $\kappa$ and percentage agreement for this setup, as well as the agreement between our expert annotators, broken down by topic.

\begin{table}\centering
\footnotesize
	\begin{tabularx}{\linewidth}{l@{\extracolsep{\fill}}ccccc}

\toprule
	  & \multicolumn{2}{c}{expert--expert} & \multicolumn{2}{c}{crowd--expert} \\
\cmidrule(r){2-3}
\cmidrule(l){4-5}
	  & \% & $\kappa$ & \% & $\kappa$ \\
\midrule
abortion 				& .884 & .651 & .834 & .660 \\
cloning 				& .845 & .712 & .821 & .704 \\
death penalty 			& .851 & .657 & .770 & .576 \\
gun control 			& .907 & .783 & .796 & .638 \\
marijuana legalizat. 	& .850 & .729 & .854 & .749 \\
minimum wage 			& .885 & .779 & .858 & .745 \\
nuclear energy 			& .809 & .686 & .889 & .825 \\
school uniforms 			& .864 & .767 & .931 & .889 \\
\midrule
average & .862 & .721 & .844 & .723 \\
\bottomrule
	\end{tabularx}	
	\caption{Agreement between experts, and between the expert and crowd gold standards.}\label{tab:agreement}
\end{table}

\begin{table*}\centering
\footnotesize
	\begin{tabular}{lrrrrr}
\toprule
	  topic & docs & sentences & no argument & support argument & oppose  argument \\
\midrule
abortion (AB)  				&  50 &  3,929 &  2,427 &   680 &   822  \\
cloning (CL)				&  50 &  3,039 &  1,494 &   706 &   839  \\
death penalty (DP) 	    	&  50 &  3,651 &  2,083 &   457 & 1,111  \\
gun control (GC) 			&  50 &  3,341 &  1,889 &   787 &   665  \\
marijuana legalization (ML) &  50 &  2,475 &  1,262 &   587 &   626  \\
minimum wage (MW)			&  50 &  2,473 &  1,346 &   576 &   551  \\
nuclear energy (NE)			&  50 &  3,576 &  2,118 &   606 &   852  \\
school uniforms (SU)			&  50 &  3,008 &  1,734 &   545 &   729  \\
\midrule
total 						& 400 & 25,492 & 14,353 & 4,944 & 6,195  \\
\bottomrule
	\end{tabular}	
	\caption{Corpus size and class distribution.}
	\label{tab:corpus_stats}
\end{table*} 

We proceeded with annotating the remaining instances in our data set using seven crowd workers each.  The workers were paid 1.2\textcent{} per instance, with each instance taking a bit less than six seconds on average.  This corresponds to the US federal minimum wage of \$7.25/hour. Our total expenditure, including AMT processing fees, was \$2,774.02.  After applying MACE with a threshold of 0.9, we were left with 25,492 gold-standard annotations.  Table~\ref{tab:corpus_stats} provides statistics on the size and class distribution of the final corpus.  The gold-standard annotations for this data set, and code for retrieving the original sentences from the Wayback Machine, are released under free licences.\footnote{\url{https://www.ukp.tu-darmstadt.de/data/}}

%------------------------------------------------
% MODELS
%------------------------------------------------
\section{Approaches for Identifying Arguments}

We model the identification of arguments as a binary, sentence-level classification and aim to learn the following function:
\begin{equation}
    \hspace*{-2mm}f(s,t) =\left\{
                \begin{array}{l}
                  0 \text{ if } s \text{ is not an argument for $t$;}\\
                  1 \text{ if } s \text{ is an argument for $t$,}\\
				\end{array}
			\right.
\end{equation}
where $s=w_1, w_2, w_3, \dots, w_n$ is a sentence consisting of words $w_i$ and $t=v_1, v_2, \dots, v_m$ is a topic with words $v_j$. 
In other words, the task is to classify sentence $s$ as ``argument'' if $s$ includes a relevant reason either supporting or opposing the given topic~$t$ and as ``no argument'' if the sentence does not include a reason or is not relevant for topic $t$.\footnote{Note that we leave stance recognition for future work.}

\subsection{Bidirectional Long Short-Term Memory Network (\model{bilstm})}
\label{sec:bilstm}

Our first model (\model{bilstm}) is a bidirectional long short-term memory network.  LSTMs~\cite{Hochreiter1997} are recurrent neural networks that process each word gradually and decide in each step which information to keep in order to produce a concise representation of the word sequence.  Traditional LSTMs, however, process the text in a single direction and do not consider contextual information of future words in the current step~\cite{tan-EtAl:2016:P16-1}.  Bidirectional LSTMs use both the previous and future context by processing the input sequence in two directions.  The final representation is the concatenation of the forward and backward step.  In order to prevent overfitting, we add dropout after the concatenation layer.  The result is fed into a dense layer with two units and softmax as the activation function.  For representing the words $w_i$, we use 300-dimensional word embeddings trained on the Google News data set by \newcite{Mikolov2013}.  To handle out-of-vocabulary (OOV) words, we create random word vectors and map each OOV word to the same random vector.\footnote{Each dimension is set to a random number between $-0.01$ and $0.01$. Digits are mapped to the same random word vector.}

\subsection{BiLSTM Model with Topic Similarity Features (\model{bilstm+cos})}
\label{sec:bilstm+cos}

A limitation of the \model{bilstm} model described in the previous section is that it does not take the topic~$t$ into account.  Consequently, the model is not able to learn the relation between sentence~$s$ and topic~$t$ and to decide if a sentence is relevant for the given topic.  To address this issue, we extend the \model{bilstm} model in the following way: we concatenate the input embedding of each word~$w_i$ with the cosine similarity between $w_i$ and the averaged word embeddings of the topic words~$v_j$. That is, we encode each word~$w_i$ of sentence~$s$ as
\begin{equation}
	\hat{x}_i = \Bigg[
                \begin{array}{c}
                  x_i\\
                  \cos(x_i, u)\\
				\end{array}
			\Bigg],
\end{equation}
where $x_i$ is the word embedding of $w_i$, $u$ is the average of the word embeddings of $v_j$ in topic~$t$, and $\cos(x_i, u)$ is the cosine similarity between $x_i$ and $u$.\footnote{We also tried concatenating $u$ with $x_i$. However, this performed worse than the vanilla \model{bilstm} model.}  We refer to this model as \model{bilstm+cos}.

\subsection{Inner-attention BiLSTM (\model{inner-att})}
\label{sec:inner-att}

In order to let the model learn which parts of the sentence are relevant (or irrelevant) to the given topic, we propose an attention-based neural network~\cite{Bahdanau2014} that learns an importance weighting of the input words depending on the given topic.  Similar approaches have been shown to achieve state-of-the-art results in aspect-based sentiment analysis~\cite{wang-liu-zhao:2016:P16-1}, question answering~\cite{tan-EtAl:2016:P16-1}, and discourse parsing~\cite{li-li-chang:2016:EMNLP2016}.
For our model, we adopt an inner-attention mechanism as proposed by \newcite{wang-liu-zhao:2016:P16-1}.
In particular, we determine the importance weighting on the input sequence instead of on the hidden states of the LSTM; this has been shown to prevent biased importance weights towards the end of a sequence. 
Following this idea, we determine the importance weighting for each input embedding~$x_i$ as
\begin{equation}
	\alpha_i = \sigma(u^T W_s x_i),
\end{equation}
where $W_s \in \mathbb{R}^{d \times d}$ are trainable parameters of the attention mechanism, $u^T$ is the transposed topic vector, and $\sigma$ is the sigmoid function to normalize the weights between 0 and 1.
Using the importance weighting, we determine the weighted input embeddings as 
\begin{equation}
	\tilde{x}_i = \alpha_i x_i
	\label{eq:weighted_embeddings}
\end{equation}
for each of the word embeddings $x_i$ of sentence $s$. 
This attention mechanism can be seen as a sieve in which uninformative words are filtered by the given topic.
For obtaining a concise representation of the sentence, we apply a BiLSTM model on the weighted input embeddings, whereas \newcite{wang-liu-zhao:2016:P16-1} used a single GRU.
Also, we do not use average pooling on the hidden layers of the RNN, but use the concatenation of the forward and backward LSTMs as the final sentence representation. 

\begin{figure}
  \centering
    \includegraphics[width=1.0\linewidth]{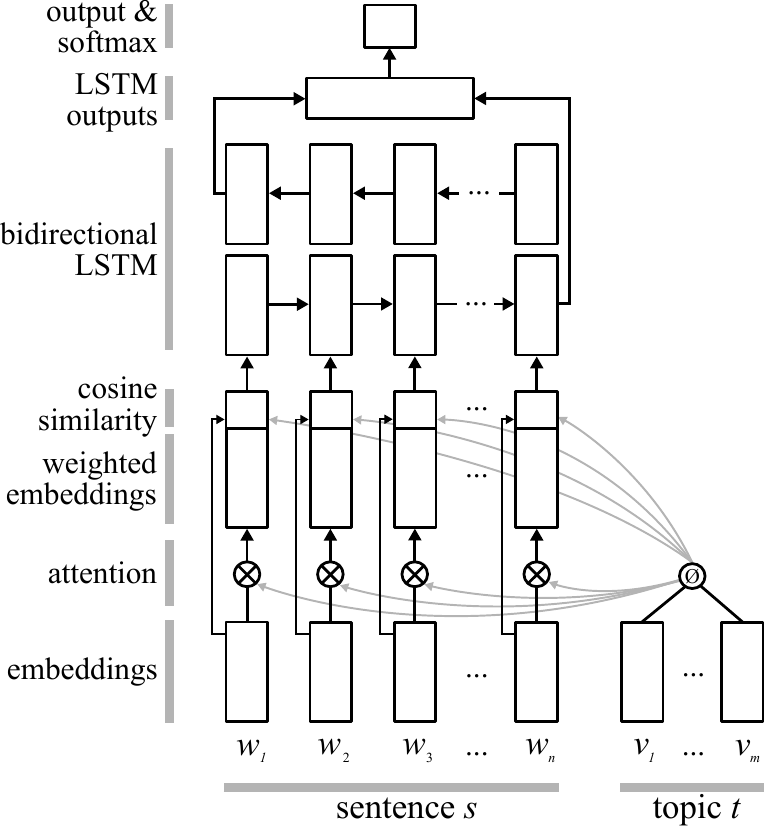}
    \caption{Schematic of the inner-attention BiLSTM model with topic similarity feature.}
    \label{fig:inner-att+cos}
\end{figure}

\subsection{Inner-attention BiLSTM with Topic Similarity Features (\model{inner-att+cos})}
\label{sec:inner-att+cos}

Our fourth model combines the inner-attention mechanism of the \model{inner-att} model and the topic similarity feature of \model{bilstm+cos}.  As with the \model{inner-att} model, we learn an importance weighting on the embeddings of the words of sentence $s$ as described in \S\ref{sec:inner-att}.  Then, we concatenate the weighted input embeddings from Equation~\ref{eq:weighted_embeddings} with the cosine similarity between the averaged topic embeddings $u$ and the embeddings of the current word $x_i$ as
\begin{equation}
	\bar{x}_i = \Bigg[
                \begin{array}{c}
                  \alpha_i  x_i\\
                  \cos(x_i, u)\\
				\end{array}
			\Bigg].
\end{equation}
Accordingly, this representation not only distills unimportant information, but also emphasizes words similar to the topic, which helps to discover off-topic sentences.
Figure~\ref{fig:inner-att+cos} shows the schematic of this model, which we refer to as \model{inner-att+cos}.

%------------------------------------------------
% EVALUATION
%------------------------------------------------
\section{Evaluation}

\begin{table*}\centering
\footnotesize
	\begin{tabular}{ ll | rrrrrrrr || rrrr }
\toprule
 && GC 	& NE 	& MW 	& AB 	& DP 	& CL 	& SU 	& ML 	& \multicolumn{4}{c}{average} \\
&& \multicolumn{8}{c||}{F\textsubscript{1}} 	& A	& F\textsubscript{1} & P & R \\
\midrule
\parbox[t]{1.5mm}{\multirow{5}{*}{\rotatebox[origin=c]{90}{in-topic}}} 
& \model{lr-uni} 		& .672	& .698	& .780	& .692	& .642	& .657	& .702	& .754 	& .702	& .700	& .649	& .728\\

\cmidrule{2-14}

& \model{bilstm} 	& .723	& \textbf{.719}	& .792	& .711	& .657	& .685	& .719	& .765	& .727	& .721	& .666	& .809\\

& \model{bilstm+cos} 	& .718	& .717	& .799	& .710	& .687	& \textbf{.740}	& .715	& .768	& .735	& .732	& .673	& \textbf{.814}\\
& \model{inner-att} 	& .732	& .712	& .816	& .725	& \textbf{.706}	& .730	& \textbf{.731}	& \textbf{.776}	& \textbf{.744}	& \textbf{.741}	& \textbf{.691}	& .800\\
& \model{inner-att+cos} & \textbf{.733}	& .706	& \textbf{.823}	& \textbf{.732}	& .702	& .722	& .716	& .752	& .739	& .736	& .682	& .793\\
\midrule

\parbox[t]{1.5mm}{\multirow{5}{*}{\rotatebox[origin=c]{90}{cross-topic}}} 
& \model{lr-uni} 		& .578	& .643	& .500	& .559	& .638	& .681	& .627	& .536	& .638	& .595	& .659	& .401\\
\cmidrule{2-14}
& \model{bilstm} 		& .681	& .577	& .435	& .542	& .676	& .711	& .526	& .586	& .655	& .592	& .728	& .376\\
& \model{bilstm+cos} 	& \textbf{.693}	& .616	& .519	& .566	& .683	& \textbf{.715}	& \textbf{.597}	& .621	& .673	& .626	& .740	& .442\\
& \model{inner-att} 	& .684	& \textbf{.618}	& .487	& \textbf{.715}	& .684	& .688	& .542	& .567	& .670	& .623	& \textbf{.749}	& .442\\
& \model{inner-att+cos} & .688	& .611	& \textbf{.684}	& .658	& \textbf{.686}	& .714	& .592	& \textbf{.632}	& \textbf{.693}	& \textbf{.658}	& \textbf{.749}	& \textbf{.516}\\
\bottomrule

	\end{tabular}	

	\caption{In-topic and cross-topic results for each of the eight topics. Bold numbers indicate the highest score in the column for in-topic and cross-topic experiments.}
	\label{tab:results}
\end{table*} 

In order to evaluate the robustness of the models, we conduct in-topic as well as cross-topic experiments. 
For the former, we use 80\% of all sentences of a topic for training and 20\% for testing. 
In order to tune the parameters of the models, we sample 10\% of the training data as validation data.\footnote{We used stratified splitting to ensure the same class distribution in all sets.} 
In cross-topic experiments, we evaluate how well the models generalize to an unknown topic.
To this end, we combine training and validation data of seven topics for training and parameter tuning, and use the test data of the eighth topic for testing. 
We intentionally do not use the entire data of the target topic for testing, since it allows us to directly compare in-topic experiments with cross-topic experiments and to investigate the influence of gradually adding target topic data to the training data (\S\ref{sec:add_target_domain_data}).

Since reporting single performance scores is insufficient to compare non-deterministic learning approaches like neural networks~\cite{reimers-gurevych:2017:EMNLP2017}, we report average scores of ten runs with different random seeds.
As evaluation measures, we report the average macro F-score over all ten runs for each topic. 
Furthermore, we report the average accuracy~(A), macro F-score, and precision~(P) and recall~(R) of the ``argument'' class over all eight topics (and runs) for in-topic and cross-topic experiments. 

We use a logistic regression model with lowercased unigram features as baseline, which has been shown to be a strong baseline for various other AM task~\cite{daxenberger-EtAl:2017:EMNLP2017,Stab2017}. We refer to this model as \model{lr-uni}.

All models are trained using the Adam optimizer~\cite{Kingma2015} and cross-entropy loss function.  For finding the best model in each of the ten runs, we stop training once the accuracy on the validation data no longer improves.  To prevent the model from overfitting, we apply dropout~\cite{Srivastava2014} for each model after the concatenation layer of the BiLSTM layer as described in \S\ref{sec:bilstm}.  To accelerate training, we set the maximum length of all sentences to $60$.\footnote{Only 244 of our sentences ($<$1\%) exceed this length.}

\subsection{Hyperparameter Tuning}

For finding the best model configurations, we tuned the hyperparamters by training each model on the training data of all topics and evaluated their performance on all validation sets. 
In particular, we experimented with LSTM sizes of 32, 48, 64, 96, 128, 160, and 192; dropouts of 0.1, 0.3, 0.5, 0.7, and 0.9; batch sizes of 16, 32, 64, and 128; and learning rates of $1 \times 10^{-2}, 5 \times 10^{-3}, 1 \times 10^{-3},$ and $1 \times 10^{-4}$.
Table~\ref{tab:hyper_parameter} shows the best parameters for each of our four models on the validation data. 
\begin{table}[!ht]\centering
	\setlength{\tabcolsep}{0.25em}
\footnotesize
	\begin{tabular}{lrrrr}
\toprule
model & LSTM size & dropout & batch size & learning rate \\
\midrule
\model{bilstm} 			& 96 	& 0.7 	& 64 	& $1 \times 10^{-2}$\\
\model{bilstm+cos} 		& 32 	& 0.7 	& 32 	& $5 \times 10^{-3}$\\
\model{inner-att} 		& 96 	& 0.3 	& 16 	& $1 \times 10^{-3}$\\
\model{inner-att+cos} 	& 128 	& 0.7 	& 16 	& $1 \times 10^{-4}$\\
\bottomrule
 \end{tabular}	
	
	\caption{Hyperparameters for each model.}\label{tab:hyper_parameter}
\end{table}

\subsection{In-topic Results}

The in-topic results (upper part of Table~\ref{tab:results}) show that all neural approaches outperform the \model{lr-uni} baseline. 
The \model{bilstm} model achieves an average accuracy of 0.727 and an F\textsubscript{1} of 0.721.
We can also see from the results that all three models using the topic as additional input perform better than the vanilla \model{bilstm} model. 
Our \model{bilstm+cos} achieves better results for four topics while the attention-based models outperform the \model{bilstm} model on seven (\model{inner-att}) and five topics (\model{inner-att+cos}). 
The \model{inner-att} model performs best in in-topic experiments, achieving  an average accuracy of 0.744 and an F\textsubscript{1} of 0.741. 
This finding suggests that the attention model successfully emphasizes those parts of the sentences which are important for the topic and that the learned importance weighting results in more concise sentence representations for AM.

\subsection{Cross-topic Results}
\label{sec:cross_domain_results}

When the target topic is unknown, the F-scores of the neural models drop on average by 0.108 (lower part of Table~\ref{tab:results}). 
In particular, all models achieve a considerably lower recall compared to in-topic experiments, which is also evident by the number of sentences classified as argument in all test sets.
For instance, in cross-topic experiments the \model{inner-att+cos} model classifies only 1,515 sentences as argument, while it recognizes 2,662 arguments in in-topic experiments. 
The results, however, also show that all neural approaches achieve a considerably higher precision compared to in-topic experiments. 
On average the precision of neural models is 0.064 better compared to in-topic experiments. 
This suggests that the neural models learn common properties that arguments  share across topics. 

The results also show that the \model{inner-att+cos} model generalizes best to unknown topics. 
It outperforms the vanilla \model{bilstm} model on all topics and achieves 0.693 accuracy and 0.658 F-score. 
The results also show that the model achieves 0.067 higher precision and the lowest drop in recall of all models compared to in-topic experiments. 
The model performs better compared to \model{bilstm+cos} and \model{inner-att}, which illustrates that the combination of the attention mechanism with the  similarity feature is helpful in cross-topic settings.

\subsection{What Does the Model Learn?}

\bgroup
\setlength{\tabcolsep}{0.25em}
\renewcommand{\arraystretch}{1.2}
\begin{table*}
  \begin{adjustbox}{max width=\textwidth}
  \begin{tabular}{c | c c c c c c c c c c c c c c c }
  \toprule
    topic & \multicolumn{15}{c}{sentence and importance weighting} \\
  \midrule
	\multirow{2}{*}{school uniforms} & \cellcolor[HTML]{cde3f2} forcing & \cellcolor[HTML]{3284bf} \textcolor{white}{students} & \cellcolor[HTML]{b1d3eb} to & \cellcolor[HTML]{3183bf} \textcolor{white}{wear} & \cellcolor[HTML]{d8e9f5} the & \cellcolor[HTML]{ffffff} same & \cellcolor[HTML]{4494cf} \textcolor{white}{clothes} & \cellcolor[HTML]{4494cf} \textcolor{white}{violates} & \cellcolor[HTML]{d8e9f5} the & \cellcolor[HTML]{3284bf} \textcolor{white}{students} & \cellcolor[HTML]{fbfdfe} right & \cellcolor[HTML]{b1d3eb} to & \cellcolor[HTML]{388dcb} \textcolor{white}{freedom} & \cellcolor[HTML]{acd0e9} of & \cellcolor[HTML]{3182bd} \textcolor{white}{expression} \\
	& \small{.048} & \small{.092}	& \small{.055} & \small{.092} & \small{.045} & \small{.034} & \small{.085} & \small{.085} & \small{.045} & \small{.092} & \small{.035} & \small{.055} & \small{.088} & \small{.057} & \small{.093} \\
	
	\midrule
	
	\multirow{2}{*}{nuclear energy} & \cellcolor[HTML]{e7f1f9} forcing & \cellcolor[HTML]{d7e8f5} students & \cellcolor[HTML]{88bbe0} to & \cellcolor[HTML]{ffffff} wear & \cellcolor[HTML]{3182bd} \textcolor{white}{the} & \cellcolor[HTML]{f3f8fc} same & \cellcolor[HTML]{feffff} clothes & \cellcolor[HTML]{aed1ea} violates & \cellcolor[HTML]{3182bd} \textcolor{white}{the} & \cellcolor[HTML]{d7e8f5} students & \cellcolor[HTML]{69aad8} \textcolor{white}{right} & \cellcolor[HTML]{88bbe0} to & \cellcolor[HTML]{ffffff} freedom & \cellcolor[HTML]{9ac5e5} of & \cellcolor[HTML]{ffffff} expression \\
	& \small{.053} & \small{.058} & \small{.078} & \small{.045} & \small{.104} & \small{.050} & \small{.047} & \small{.068} & \small{.104} & \small{.058} & \small{.087} & \small{.078} & \small{.047} & \small{.074} & \small{.047} \\

  \bottomrule
  \end{tabular}
  \end{adjustbox}	
  \caption{Influence of topic relevance (first row) and irrelevance (second row) on attention weighting.}
  \label{tab:vis_attention_weights}
\end{table*}
\egroup

In an attempt to understand what the attention-based model learns, we analyzed the importance weights of individual words. 
Table~\ref{tab:vis_attention_weights} shows how the importance weighting of the \model{inner-att+cos} model changes for the same sentence when different topics are given as input. 
The first row shows the importance weights for the topic ``school uniforms'', to which the sentence is relevant.
As the colours indicate, the model gives high attention to words like ``students'' and ``wear'', which are relevant to the topic. More importantly, the model also emphasizes words like ``violates'', ``freedom'', and ``expression'', which represent the gist of the argument. 
The second row shows the importance weighting when providing a topic not relevant to the sentence. 
As we can see, the model gives higher attention to stop words like ``the'', ``of'', and ``to''.
It also gives attention to words like ``violates'' and ``right'' which are less topic-dependent and likely to appear in arguments relevant to other topics. 
This example illustrates that our attention-based model successfully learns which words make a sentence a relevant argument for a given topic. 

As the evaluation results suggest, our model learns specific features allowing it to achieve high precision in cross-topic experiments (see \S\ref{sec:cross_domain_results}).
In order to better understand these features, we ranked the words of all positively classified sentences in the test sets according to their average importance weights in all cross-topic experiments. 
Among the top-ranked words are remarkably few topic-dependent words, but many adverbs and adjectives like ``fair'', ``perfect'', ``wrong'', ``easier'', and ``impossible''.  
Also, verbs like ``infringe'', ``oppose'', and ``undermine'' receive high importance weights across topics. 
This shows that the attention-based model gives high attention to words that assign positive or negative attributes to specific entities.

\subsection{Error Analysis}

To better understand the errors of the \model{inner-att+cos} model, we manually analyzed 100 sentences randomly sampled from the false positives and false negatives of cross-topic experiments.  Among the false positives, we found 42 off-topic sentences that were wrongly classified as arguments.  The 58 on-topic false positives are primarily non-argumentative background information about the topic, or mere opinions about the topic without evidence (cf.~the first and fourth examples in Table~\ref{tab:examples}).  Among the false negatives, we found 61 sentences not explicitly referring to the topic but to related aspects that make the sentence a relevant argument.  For instance, the model fails to establish argumentative links between the topic ``cloning'' and aspects like diminishing the waiting lists for organ donation, or links between ``nuclear energy'' and the conditions of workers in uranium mines.

\subsection{Adapting to New Topics}
\label{sec:add_target_domain_data}

\begin{figure}
    \centering
    \subfigure{\includegraphics[width=0.49\linewidth]{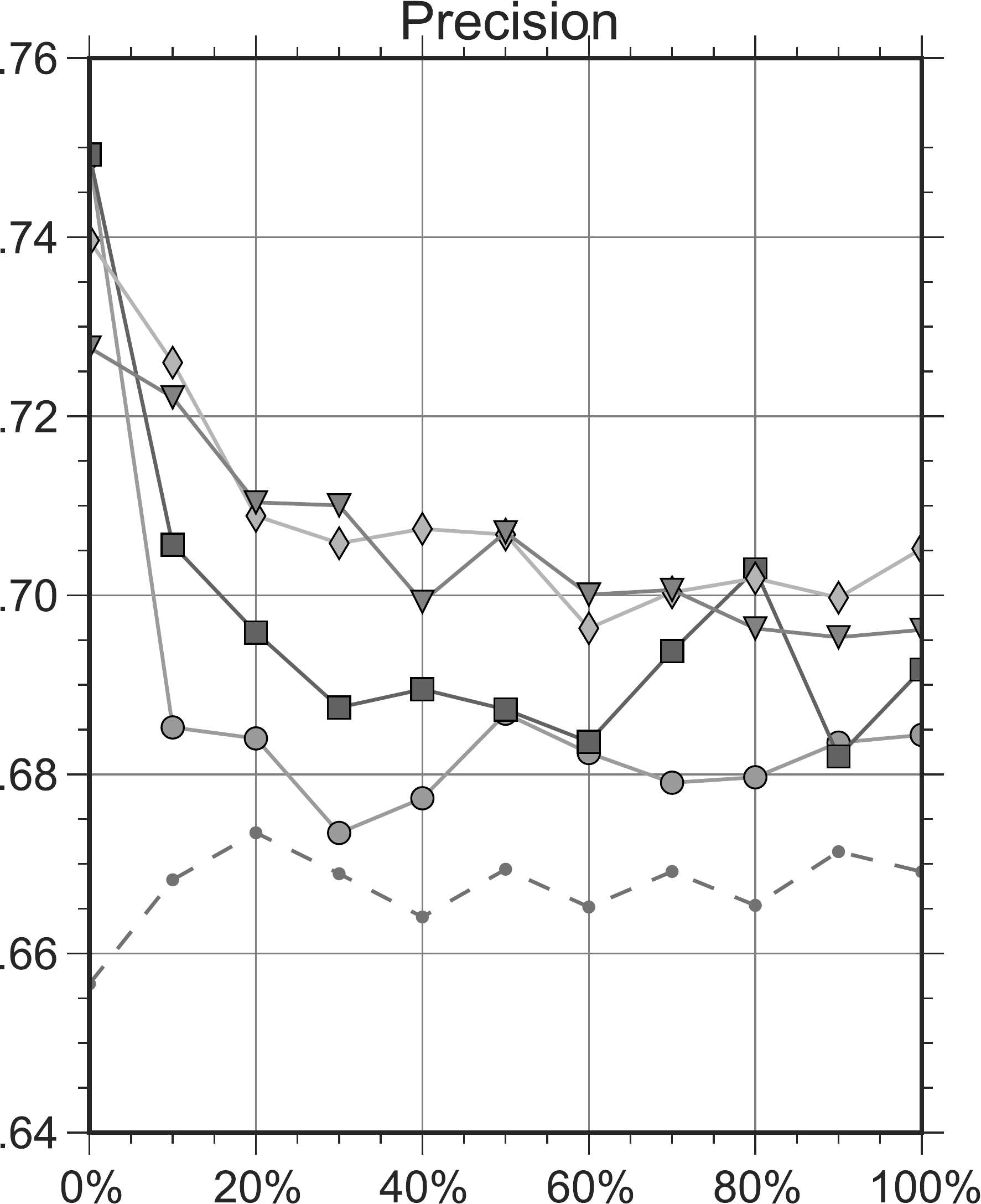}}
\hfill
    \subfigure{\includegraphics[width=0.49\linewidth]{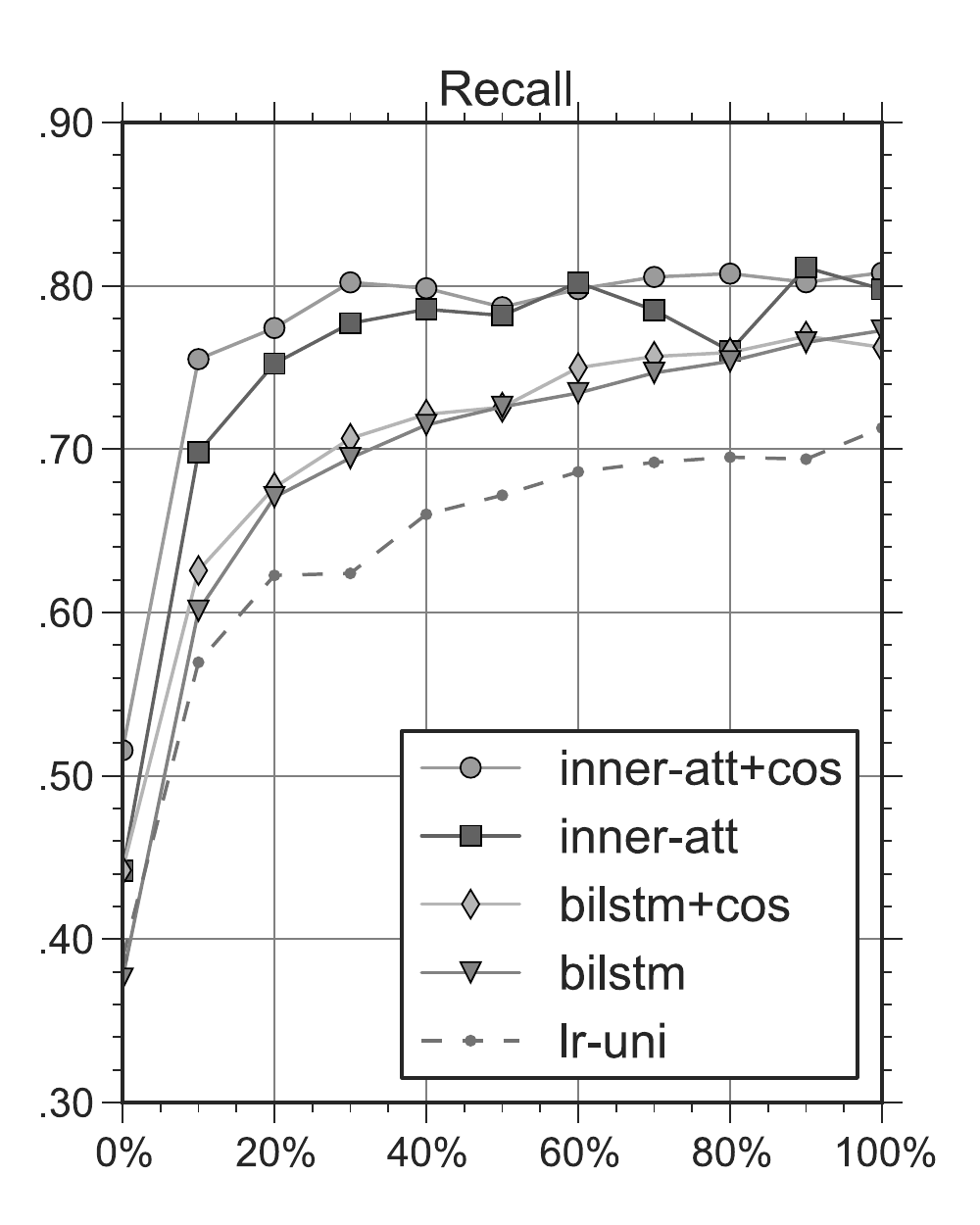}}
\caption{Model performance ($y$-axes) according to the amount of target-topic data in the train sets ($x$-axes).}
\label{fig:extend_train_data}
\end{figure}

In order to quantify the amount of topic-specific data required by the models to achieve in-topic results, we gradually add target topic data in cross-topic experiments to the training data and evaluate model performance on the target test set. 
Figure~\ref{fig:extend_train_data} shows the average precision and recall over all topics when adding different amounts of randomly sampled topic specific data to the training data ($x$-axes).\footnote{Each data point in the plot is the average score of $80$ experiments (ten runs with different random samples of target-topic data for each of the eight topics).}
As the results show, all models achieve higher recall, while the precision drops when adding target topic data to the training data.
This shows that the models tend to emphasize topic-related information more than topic-independent features when target-topic data is available. 

This effect is most evident for the \model{inner-att+cos} model, which uses information about the topic in the attention mechanism as well as in the similarity feature. 
The results, however, also show that the \model{inner-att+cos} model achieves 0.802 recall with only 30\% of the target topic data, while the vanilla \model{bilstm} model and \model{bilstm+cos} model do not reach in-topic recall with all available target topic data.

%------------------------------------------------
% CONCLUSION
%------------------------------------------------
\section{Conclusion}

We have presented a new approach for searching a document collection for arguments relevant to a given topic.  First, we introduced an annotation scheme that is applicable to the information-seeking perspective of argument search and general enough for use on heterogeneous texts.  Second, by comparing crowdsourced annotations to expert annotations, we showed that our annotation scheme is reliably applicable by untrained annotators to arbitrary Web texts.  Third, we presented a new corpus, including over 25,000 instances over eight topics, that allows for cross-topic experiments using heterogeneous text types.  The annotations as well as the source code for downloading the sentences from the Wayback Machine are made available for future work.  Fourth, we conducted in- and cross-topic experiments and showed that our attention-based model better generalizes to unknown topics than vanilla BiLSTM models.

\bibliography{bibliography}
\bibliographystyle{acl_natbib}

\end{document}